\documentclass[conference]{IEEEtran}
\IEEEoverridecommandlockouts
\usepackage{cite}
\usepackage{amsmath,amssymb,amsfonts}
\usepackage{graphicx}
\usepackage{textcomp}
\usepackage{xcolor}

\usepackage{balance} 
\usepackage[linesnumbered,ruled,vlined]{algorithm2e}
\SetKwInOut{End}{} 
\usepackage{caption}
\usepackage{subcaption}
\usepackage{booktabs}
\usepackage[left=0.75in, right=0.75in, top=0.75in, bottom=0.75in]{geometry}
\usepackage{soul}
\usepackage{copyright}

\def\BibTeX{{\rm B\kern-.05em{\sc i\kern-.025em b}\kern-.08em
    T\kern-.1667em\lower.7ex\hbox{E}\kern-.125emX}}

\newgeometry{left=0.75in, right=0.75in, top=1in, bottom=0.75in}

\begin{document}

\restoregeometry

\title{Generating and Explaining Corner Cases Using Learnt Probabilistic Lane Graphs
}

\author{\IEEEauthorblockN{1\textsuperscript{st} Enrik Maci}
\IEEEauthorblockA{\textit{Dept. of Engineering Science} \\
\textit{University of Oxford}\\
enrik.maci@wadham.ox.ac.uk}
\and
\IEEEauthorblockN{2\textsuperscript{nd} Rhys Howard}
\IEEEauthorblockA{\textit{Oxford Robotics Institute} \\
\textit{University of Oxford}\\
rhyshoward@live.com}
\and
\IEEEauthorblockN{3\textsuperscript{rd} Lars Kunze}
\IEEEauthorblockA{\textit{Oxford Robotics Institute} \\
\textit{University of Oxford}\\
lars@robots.ox.ac.uk}
\thanks{
This work was supported by EPSRC project RAILS (grant reference: EP/W011344/1), and Oxford Robotics Institute research project RobotCycle.}}

\maketitle

\begin{abstract}
Validating the safety of Autonomous Vehicles (AVs) operating in open-ended, dynamic environments is challenging as vehicles will eventually encounter safety-critical situations for which there is not representative training data. By increasing the coverage of different road and traffic conditions and by including corner cases in simulation-based scenario testing, the safety of AVs can be improved. However, the creation of corner case scenarios including multiple agents is non-trivial. Our approach allows engineers to generate novel, realistic corner cases based on historic traffic data and to explain why situations were safety-critical. 

In this paper, we introduce Probabilistic Lane Graphs (PLGs) to describe a finite set of lane positions and directions in which vehicles might travel. The structure of PLGs is learnt directly from spatio-temporal traffic data. The graph model represents the actions of the drivers in response to a given state in the form of a probabilistic policy. We use reinforcement learning techniques to modify this policy and to generate realistic and explainable corner case scenarios which can be used for assessing the safety of AVs.
\end{abstract}


\copyrightnotice

\section{Introduction}
\noindent In real-world driving scenarios, a driver will often interact with and consider multiple background vehicles at each instance in time. For each background vehicle, the driver must consider their position, speed, and future trajectory relative to their own. Each of these features introduces separate dimensions of variability, therefore, attempting to emulate such decision-making in simulation results in high-dimensional models which lack explainability.

We introduce the PLG that describes the spatial layout of the road network under consideration. The graph defines a discrete set of locations in which vehicles may exist, with edges defining the directions in which vehicles may travel. This framework is then used to build low-dimensional statistical models which output vehicle actions given the vehicle position and kinematic state.

\begin{figure}[t]
  \centering
  \fbox{\includegraphics[width=0.75\columnwidth, clip, trim=10em 12em 8em 10em]{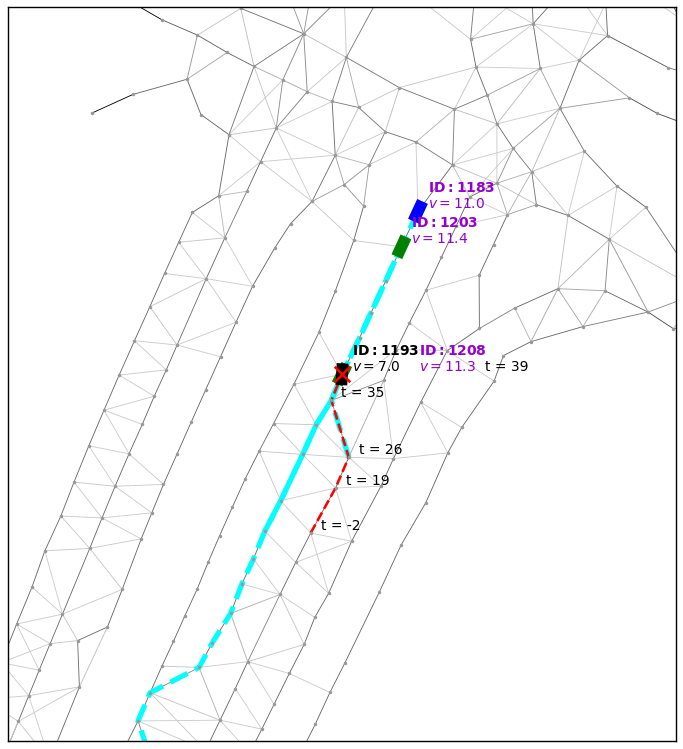}}
  \caption{A corner case generated using PLG learnt from real-world observations. In this example, the black vehicle (ID 1193) attempted a sharp cut-in in front of another vehicle (ID 1208) leading to a crash event (at $t=39$). Due to the higher velocity of the trailing vehicle a rear-end collision could not be prevented. Note that vehicle 1193 (black) is plotted on top of vehicle 1208.}
  \label{fig:cc1}
\end{figure}

Producing intelligent human-like behaviour in simulation often requires very high-dimensional models which lack explainability and behave much like a black box. high-dimensional models are often approximated by deep neural networks (DNNs) which further increase the cost of simulation due to their requiring GPU technology or long periods of time to train appropriately.

In this paper, new corner cases are generated in an entirely simulated environment in the form of logged spatio-temporal events. The proposed graphical model allows for high volumes of realistic timeline data to be generated and provides highly understandable explanations. Data is simulated from an initial state up until a corner case event (e.g. a crash or a near-miss) between a simulated autonomous vehicle (AV) and background vehicle (BV) occurs. Ultimately, corner case events are generated using this novel methodology along with visualisations that can provide intuitive, human-understandable explanations of the vehicle behaviour policy.
To this end, the paper makes the following contributions:
\begin{itemize}
    \item We introduce Probabilistic Lane Graphs (PLGs) to represent and visualise highly explainable and human-understandable corner case scenarios accounting for multiple agents;
    \item We present algorithms to learn PLGs from real-world traffic data; and
    \item We describe a novel framework for corner case generation using reinforcement learning on PLGs.
\end{itemize}

Fig.~\ref{fig:cc1} depicts an example generated corner case scenario terminating with a rear-end collision.
The paths of each of the vehicles are highlighted in the figure. This scenario is generated by the proposed method in this paper. The PLG is used to discretise and generate vehicle trajectories whilst a separate model is trained to react to BVs and generate vehicle actions.


\section{Related Work}
\label{sec:relwork}
In the context of AV testing and validation, the generation of corner cases is an important and active area of research. Corner cases include scenarios that are caused by e.g. new object types, an anomalous traffic pattern, or an unusual combination of objects, as described by \cite{lit:cc_definition_0}, which ultimately leads to a crash event between two vehicles. For machine learning, corner cases are important as they are required for training, verification, and improved performance of ML models during inference within automated driving systems \cite{lit:cc_definition}. Most works distinguish between perception and decision-based corner cases. Here we focus on the latter. Generating corner cases in simulation is critical because to safely validate an AV’s driving system would require hundreds of billions of natural driving data points \cite{lit:cc_motivation}. Decision-based corner cases usually correspond to a sequence of events (or states) terminating with a crash event or high-risk incident due to decisions made by an AV and/or surrounding agents within its environment.

In \cite{lit:learning_to_collide}, an adaptive framework for safety-critical scenario generation is proposed by dividing the traffic scenarios into some reusable building blocks and building dependency among them with human knowledge. These building blocks form a joint policy which is then optimised by rewarding it when risky scenarios are generated. The work, however, does not include scenarios where the AV interacts with multiple agents and is limited to simple scenarios.

Work by \cite{lit:risky_index} generates safety-critical test scenarios by sampling parameters from a probabilistic model based on naturalistic driving data which consists of vehicle kinetic information and a traffic risk index. The proposed method allows for sampling parameters for mathematical functions that represent the vehicle kinematics during a manoeuvre. The work is mainly restricted to modelling cut-in scenarios.

On the other hand, \cite{lit:AST_likely_path} is not only concerned about whether a failure can occur but also discovering which failures are most likely to occur. The article uses adaptive stress testing and considers the simulator as a partially observable black box setting with continuous-valued systems operating in an environment with stochastic disturbances.

In \cite{lit:efficient_stat_val}, reinforcement learning is used to learn the behaviours of simulated actors that cause unsafe behaviour measured by the well-established Responsibility-Sensitive Safety (RSS) metric. By biasing automatically generated test cases towards the worst-case scenarios potential unsafe edge cases can be identified. However, in this case, the action chosen for the pedestrian is very simple.

Work by \cite{lit:corner_case_gen} addresses some of the limitations of the existing decision-making corner case generation method. Existing methods are only validated under simplified scenarios and cannot process highly complex environments. As stated in \cite{lit:learning_to_collide}, finding the proper representation is a difficult task, since scenarios consisting of statistic and dynamic objects, routes, and maps, are hard to model. 
In \cite{lit:corner_case_gen}, scenarios are modelled as interactions between multiple traffic participants, which can handle the curse of dimensionality in space. Value-based deep reinforcement learning, also known as the Deep Q Network, is then used to learn the optimal policy to allow BVs to interact more aggressively with the AV.

To simulate real-life scenarios more closely, the dimensionality of the model must increase to account for the wide range of variables a human driver must consider. However, the increase in dimensionality ultimately results in a reduced means of model explainability. Deep neural networks are commonly used to approximate complex policies, however, at present, there is still a lack of insight into the internal logic and rationale behind DNN predictions --- they behave much like a black box. In \cite{lit:explainability} it is stated that explainability of an AV's decision-making system involves the provision of insights into the planning operations (behaviour planning and path planning) of the AV. In this work, we introduce the PLG framework which allows us to run the path planning and action generation separately and in parallel, providing a high degree of explainability leading up to the final state of the simulated cases. As a result, we can produce corner case events and study the simulated driver actions leading up to the safety-critical event. A similar approach to our PLG framework is explored in \cite{lit:lane_graph}, where lane graphs are extracted from a bird's eye view image of a road network. This is because lane-level scene annotations provide invaluable data for AVs for trajectory planning in complex environments such as urban areas and cities. On top of this, a neural network, \emph{LaneDirNet}, is used to predict the directions of lanes. However, in this work, we extract the PLG using timeline data and we work with graphs on a much larger scale. Furthermore, the lane directions are inherent in the learnt adjacency matrix of the PLG. Closer to our PLG structure is the work by \cite{lit:graph_representation}. This work aims to develop a formal method to model high-level interaction between autonomous vehicles by using \emph{road graphs} to represent complex road situations. Their definition of a road graph is comparable to our definition of a PLG. On top of this, they also define traffic control protocols which describe the flow of traffic through the road graph with respect to time. The road graphs and time protocols are not utilised further beyond the introduction of their definitions. We will show how these definitions and time protocols can be utilised in conjunction with our PLG to aid the explainability of an agent's trajectory and actions. 

In the field of path planning, the work in \cite{lit:trajectory_planning} creates a path toward the desired goal and then tracks this path by generating a set of candidate trajectories that follow the path to varying degrees and selecting from this set the best trajectory according to an evaluation function. We will show that we can achieve a similar outcome via a significantly simplified method due to our discretisation of the state space. The work presented by \cite{lit:scene_graphs} uses \emph{scene graphs} to provide scene understanding and improve the explainability of their convolutional neural network for risk assessment of driving actions. The work primarily focuses on scene graph extraction from image and video data and can complement our work on PLG generation.

The state of the art in this field includes a variety of longitudinal (cruise control/car following) microscopic models accounting for regular vehicles, AVs and Connected AVs (CAVs). G\'asp\'ar et al. \cite{lit:soa_acc} proposes an adaptive cruise control (ACC) speed design method for road vehicles, which minimizes control energy and fuel consumption while keeping travelling time and, moreover, considers the local traffic information to avoid conflicts in congestions. The work used vehicle-to-vehicle and vehicle-to-infrastructure communications to adapt vehicle speeds accordingly.

Chin et al. \cite{lit:soa_cacc} presents cooperative adaptive cruising control (CACC) of multiple cars in automated/un-automated mixed traffic. The PrARX model is used to account for un-automated car vehicle manoeuvre's, which is a continuous approximation of hybrid dynamical system. The acceleration inputs of automated vehicles are computed in model predictive control framework where the state equation includes a platoon of automated and un-automated cars coupled with PrARX driver models.

Xiao et al. \cite{lit:soa_car_following} developed a realistic and collision-free car-following model for ACC-CACC vehicles. A multiregime model combining a realistic ACC-CACC system with driver intervention for vehicle longitudinal motions is proposed. The model accounts for driver intervention and tested three regular scenarios of stop-and-go, approaching, and cut-out maneuvers, as well as two extreme safety concerned maneuvers of hard brake and cut-in.

Li and Ma \cite{lit:soa_cffm} propose a Collision-Free car-Following Model (CFFM) for connected automated vehicles that can prevent rear-end collisions when the preceding vehicle suddenly brakes. The first step is to obtain an acceleration set that can ensure safety under predictable dangerous conditions, and then to decide on the exact acceleration by considering efficiency and comfort. It is demonstrated that CFFM has outstanding performance in terms of both efficiency and string stability.

\section{Probabilistic Lane Graphs}
\label{sec:plg}

In this section we define PLGs, explain the pipeline used to extract them from spatio-temporal traffic data, and describe how such a framework can help to facilitate explainability of an autonomous driving model.

\subsection{Definition}

Given a set of traffic data $\mathcal{D}$, a PLG is defined as follows:
a \emph{PLG} is a directed graph $\mathcal{G}$ composed of a set of nodes  $\mathcal{N}$ and edges (described in an adjacency matrix $\mathcal{A}$)  where each node in the graph describes a discrete position in space in which a vehicle has been observed ($\mathcal{D}$), and, each directed edge describes how a vehicle may navigate between nodes.
The graph, when extracted from $\mathcal{D}$, accurately recreates the road network for studying vehicle behavior. Furthermore, all edges include probabilities determined by the observed transition frequencies in $\mathcal{D}$.  

\begin{figure}
  \centering
   \includegraphics[width=0.34\linewidth]{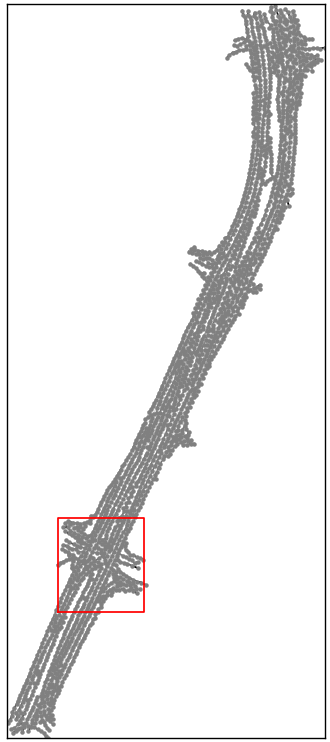}\hfill
  \includegraphics[width=0.651\linewidth]{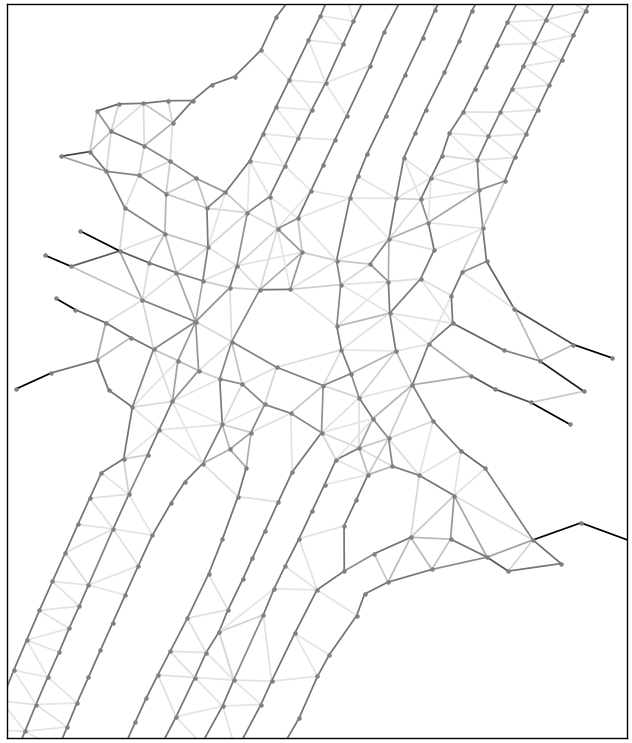}
  \caption{Probabilistic Lane Graph (PLG) for real-world highway data \cite{dataset:NGSIM}. \emph{Left:} Entire PLG. \emph{Right:} Zoomed-in subset of the PLG representing vehicle positions and possible transitions, with edge shading indicating the probability of a vehicle traversing an edge. }
  \label{fig:PLG}
\end{figure}

Graphs have long been used as a powerful tool for studying a set of interrelated objects (the nodes in the graph). Furthermore, researchers in the field of robotics have been known to apply graph discretisations to solve a range of robotic path-planning problems. However, such solutions have often suffered from the curse of dimensionality when the graph discretisations are large. In this work, we address the path planning problem on large graphs (in the order of thousands of nodes) using an efficient data-driven approach whilst maintaining a precise discretisation of the road network. An example PLG extracted from a data set is shown in Fig.~\ref{fig:PLG}. The edges are shaded with their respective probabilities, i.e., edges that were accessed more often are shaded darker. We can distinctly notice the formation of a set of lanes and how these lanes merge into the intersection. Furthermore, looking at the edges between adjacent lanes we can notice where lane transitions were most common within the road network. Such information gives the designer option to hone in on particular regions where a particular manoeuvre was most common to study vehicle behaviour and increase the probability of the given event. In this work, we fully describe the PLG, $\mathcal{G}$, using a set of coordinates for the nodes, $\mathcal{N}$, and an adjacency matrix of probabilities, $\mathcal{A}$, hence, $\mathcal{G} = \left(\mathcal{N}, \mathcal{A}\right)$. 

As well as providing an intuitive visualisation of the spatial layout of the road network and the general paths taken by vehicles in $\mathcal{D}$, we also exploit the PLG to construct a more explainable, human-understandable driver model. We utilise the PLG to break down the driver model into two distinct parts: \textit{path generation} and \textit{action generation}. Existing work in this field utilises a single model to simulate driver behaviour. The resulting parallel framework can simulate more complex environments using low-dimensional models, hence, increasing model understandability.

Commonly employed graph path planning algorithms, such as Dijkstra and A*, would be too slow to converge given our large number of nodes. In the following, we will first describe a pipeline used to extract the PLG from a spatio-temporal traffic data $\mathcal{D}$, and then introduce our novel path-planning algorithm and demonstrate how this facilitates explainability.

\subsection{PLG Generation}
Generating the PLG requires two main steps: generation of the set of nodes, $\mathcal{N}$, and generation of a set of node connections and their associated probabilities, $\mathcal{A}$.

Algorithm \ref{algorithm:node generation} is used to extract a set of nodes, $\mathcal{N}$, from the traffic data. The symbol definitions are as follows: $R$ is the minimum node spacing in the PLG, $N_v$ is the total number of vehicle paths, $\mathcal{D}_{v=i}$, is the subset of spatio-temporal data describing the kinematics of the $i$th vehicle, $\mathbf{x}$ is a 2D spatial coordinate describing a vehicle position and, $\mathbf{x}_\mathcal{N}$ is used to denote the coordinates of the nodes in the PLG. The $k$-means step is applied to the node set to create a smoother distribution of nodes along each lane. The subscript $l$ indicates lane ID, $N_l$ is the total number of lanes, and, $\mathcal{D}_l$ is the subset of data corresponding to lane $l$. We perform $N_l$ different $k$-means fits where the number of cluster centres, $k$, for each fit is the number of nodes in the PLG corresponding to that lane ID, $l$. We constrain $R$ via a qualitative lower and upper bound; $R$ must be large enough such that we do not get nodes that are laterally adjacent to each other within the same lane and, small enough such that the PLG captures the details of the road network. For the example given in Fig.~\ref{fig:PLG} $R$ was in the region of 2--3 $m$. One drawback here is the lack of a quantitatively optimal method to calculate $R$.

\begin{algorithm}[tb]
\caption{Node Set, $\mathcal{N}$, Generation Algorithm}
\label{algorithm:node generation}
\SetKwData{Left}{left}\SetKwData{This}{this}\SetKwData{Up}{up}
\SetKwFunction{Union}{Union}\SetKwFunction{FindCompress}{FindCompress}
\SetKwInOut{Input}{Input}\SetKwInOut{Output}{Output}
\Input{The original data set $\mathcal{D}$}
\Output{A set of nodes $\mathcal{N}$ for the PLG}

\texttt{\color{teal} \% Generate an initial set of nodes \color{black}}

Initialise the set of nodes in the PLG as $\mathcal{N} = \emptyset$

\For{$i = 0$ to $i = N_v-1$}{
    \For{ $\mathbf{x}$ in $\mathcal{D}_{v=i}$}{
        \If{$\mathcal{N}$ == $\emptyset$ \textbf{or} $||\mathbf{x}_\mathcal{N} - \mathbf{x}||_2 > R$ for every node in $\mathcal{N}$}{
            $\mathcal{N} = \mathcal{N}~\bigcup~\mathbf{x}$
        }
    }
}

\texttt{\color{teal} \% Even out lane node distribution \color{black}}

\For{$i = 0$ to $i = N_l-1$}{
    Perform a k-means fit on the subset of data $\mathcal{D}_{l=i}$ with $\mathcal{N}_{l=i}$ as the initial centres to return $\mathcal{N}_{l=i}^*$
}

\end{algorithm}

To generate $\mathcal{A}$, we convert the spatial path of each vehicle into a list of nodes by approximating its position at any given time by the nearest node at that time, i.e., a nodal path through the PLG. The nodal path is then converted into a unique set of nodes by removing any duplicate node entries whilst maintaining the chronological order of the nodes in the list. Duplicates may be present if, for example, a vehicle was stationary at a traffic light for a set period. An edge is then added to the PLG between any two nodes which appear adjacent to each other in the vehicle path described by the unique set of nodes. The probability distribution of the vehicle behaviour is learnt using a frequentist approach based on the observations in the data set. If a vehicle within $\mathcal{D}$ traversed the edge $i \rightarrow j$ we increment the matrix element $\mathcal{A}[i,j]$ by 1.

\subsection{Vehicle Path Planning}
The path planning algorithm is one of the two parallel models in our driver behaviour simulator. Given that, a driver knows their current location and where they would like to go, they can plan their future path in the absence of BV traffic. In the context of the PLG framework, given that we know our current node and our target node we generate the next node in the vehicle path. The probabilistic formulation of this is:
\begin{equation} \label{eq:path plan}
    p(n_{k+1}~|~n_k, c_i)
\end{equation}
where $n_{k+1}$, $n_k$, and $c_i$ are the next node, current node, and target destination. We note that the target destination, $c_i$, is not a node, but rather a cluster of target nodes. We terminate our path generation once the vehicle reaches any node within this target cluster. Mathematically, $c_i$ is a set of nodes in the PLG corresponding to the $i$th exit in the map. Furthermore, we also define the list $\mathbf{C}_i$ to be a list of target clusters ordered by which is closest to target cluster $c_i$, where the list $\mathbf{C}_i$ will include $c_i$ itself. For example, if a vehicle's target cluster destination is $c_{i = 5}$ and $c_{i=3}$ is the next closest target cluster, in Euclidean distance, then the list would be $\mathbf{C}_{i=5} = [c_{i=5}, c_{i=3}, ...]$ since the two closest target clusters to $c_{i = 5}$ are itself and then $c_{i = 3}$.

Algorithm \ref{algorithm:path generation} is then used to generate vehicle paths. We continually attempt to sample from (\ref{eq:path plan}) until we have found a viable next node. Initially, we attempt to sample from the target cluster we would like to find a path to, however, given that the data set is limited, there may be a pair of conditionals, $\{n_k, c_i\}$, that we encounter in simulation that may not have appeared in the data set, $\mathcal{D}$. Since (\ref{eq:path plan}) is learnt directly from the data set, this would mean we cannot generate a next node for these conditionals. In this case, we change the conditional to the next closest cluster and attempt to sample from this. We repeat this process until we find a viable next node. Sampling from the closest clusters to our target guides the vehicle in the general direction of its target until it can find a valid set of conditionals, $\{n_k, c_i\}$, for its actual target destination. When there is no viable path to the desired $c$, the vehicle ends up at the most likely exit from its current position.

\begin{algorithm}[tb]
\caption{Path Planning Algorithm}
\label{algorithm:path generation}
\SetKwData{Left}{left}\SetKwData{This}{this}\SetKwData{Up}{up}
\SetKwFunction{Union}{Union}\SetKwFunction{FindCompress}{FindCompress}
\SetKwInOut{Input}{Input}\SetKwInOut{Output}{Output}
\Input{A starting position $n_k$, and a target destination $c_i$}
\Output{A path from $n_k$ to $c_i$ as a list of nodes}

Initialise a starting node $n_0$ at time $k=0$ and target cluster $c_i$

\While{$n_k$ not in $c_i$}{
    \For{$c$ in $\mathbf{C}_i$}{
        \If{$\{n_k, c\} \in \mathcal{D}$}{
            sample $n_{k+1} \sim p(n_{k+1}~|~n_k, c)$
            
            $n_k = n_{k+1}$
            
            Break out of for loop
        }
    }
}
\end{algorithm}

Similar to the trajectory generation described in \cite{lit:trajectory_planning}, one can imagine using (\ref{eq:path plan}) to generate a tree of possible different paths that the vehicle could have taken which all converge to $c_{i = 5}$. Furthermore, given that we have direct access to and the ability to compute (\ref{eq:path plan}) efficiently, we can directly compute the probability of each path. However, in our case, the process is simplified significantly due to our space discretisation via the PLG. Hence, we provide the visualisation and the capability to simulate each of the various paths the AV can take to reach its target destination, in addition to, calculating the likelihood of each path.

The Path Planning Algorithm is capable of generating paths that follow the rules of the map with no extra encoding. For example, the algorithm will not generate a path that travels south in a set of nodes corresponding to a lane structure in which vehicles may only travel north because the adjacency matrix is directed. Both of the algorithms described here are made available via the GitHub repository in \cite{GitHub:plg-generation}.

\section{Corner Case Generation}
\label{sec:cornercase}

We define a corner case in the context of the PLG framework as follows: a \emph{Corner Case} is a simulated sequence of states that terminates with a specific event (e.g. a crash or near-miss) between two vehicles. Such an event occurs if two vehicles exist on the same node simultaneously. 

In the following section, we describe how we generate corner cases using PLGs.  
\begin{figure*}[t]
    \centering
    \begin{subfigure}{0.3\linewidth}
        \centering
        \fbox{\includegraphics[width=\linewidth, clip, trim=17em 21.5em 3em 3.5em]{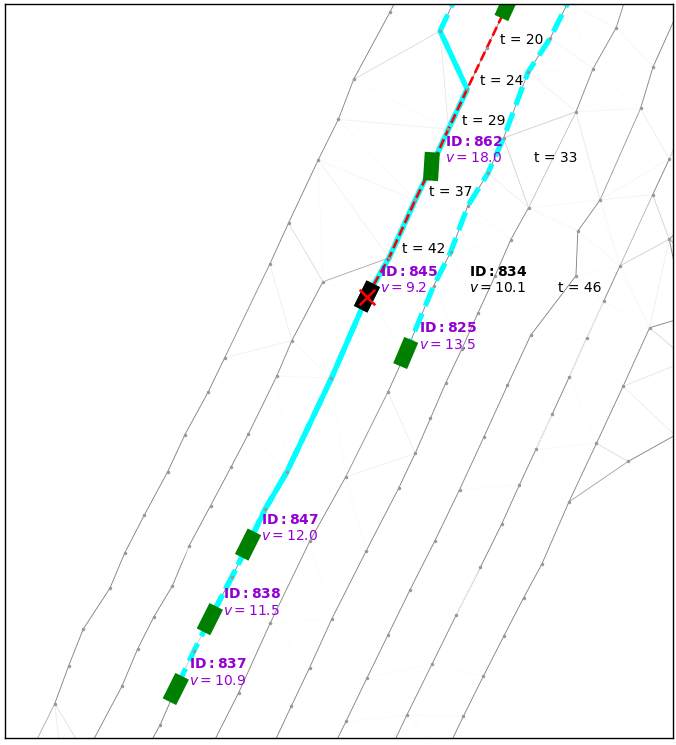}}
        \subcaption{\textit{Case 1:} no lane changes.}
        \label{subfig:cc_case_1}
    \end{subfigure}
    \hspace{0.025\linewidth}
    \begin{subfigure}{0.299\linewidth}
        \centering
        \fbox{\includegraphics[width=\linewidth, clip, trim=13em 19em 7em 6em]{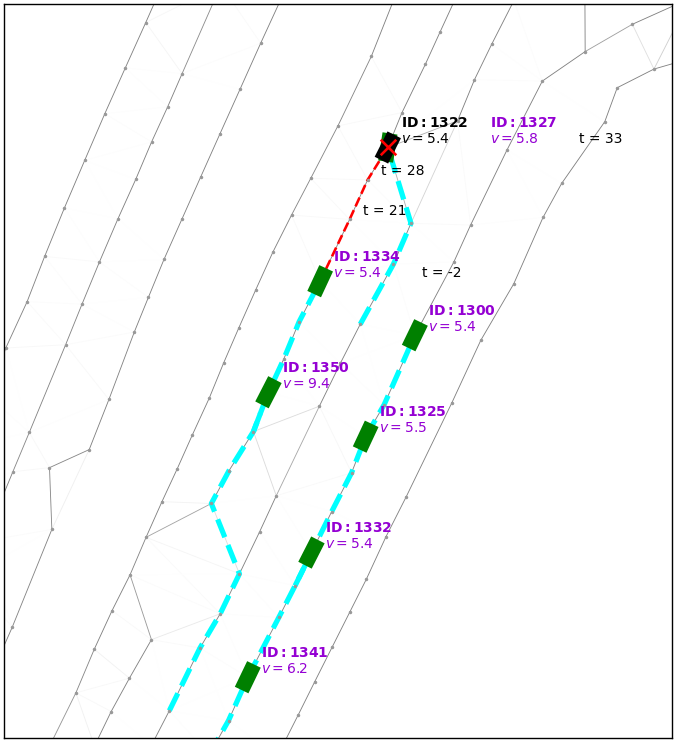}}
        \subcaption{\textit{Case 2:} one lane change.}
        \label{subfig:cc_case_2}
    \end{subfigure}
    \hspace{0.025\linewidth}
    \begin{subfigure}{0.305\linewidth}
        \centering
        \fbox{\includegraphics[width=\linewidth, clip, trim=5em 0.5em 15em 24.5em]{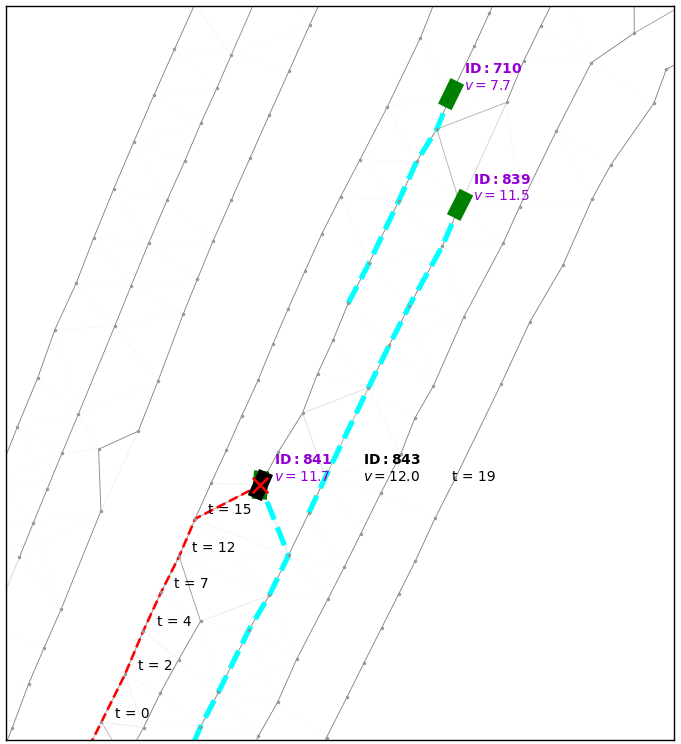}}
        \subcaption{\textit{Case 3:} two (or more) lane changes.}
        \label{subfig:cc_case_3}
    \end{subfigure}
    \caption{Corner case scenarios leading to accidents with different numbers of lane changes.}
    \label{fig:cc cases}
\end{figure*}
To this end, we propose a complete system to simulate driver actions in a multi-agent environment by combining the two parallel models with the PLG. We generate vehicle actions in the form:
\begin{equation}
    p(action~|~state) \label{eq:action general}
\end{equation}
where we define vehicle actions, at time step $k$, to be accelerations and lane changes:
\begin{equation}
    action = \{a_k, L_k\}
\end{equation}
and the state at a given time is described by the risk imposed upon the AV by the surrounding BVs:
\begin{equation}
    state = \mathbf{r}_k
\end{equation}
Hence, the overall action generation distribution becomes:
\begin{equation} \label{eq:action}
    p(a_k, L_k~|~\mathbf{r}_k)
\end{equation}

The separation of the path planning problem into an independent model means that we only need to generate the vehicle kinematics in (\ref{eq:action general}). Acceleration and lane change outputs allow for a wide range of complex vehicle actions to be simulated. The variable $a$ controls the vehicle kinematics, whilst $L$ can be used to force a lane change when desired, e.g., in an overtaking scenario. Furthermore, in the conditional, we need only consider the BV kinematics (which is encompassed within the risk variable) since all vehicle trajectories and their associated probabilities can be fully described by (\ref{eq:path plan}).

The conditional in (\ref{eq:action}) describes the risk imposed upon the AV due to the BVs, if there are $m$ total BVs we have:
\begin{equation}
    \mathbf{r}_k = \{r_{k,0}, r_{k,1}, ..., r_{k,m}\}
\end{equation}
We use the Modified Time to Collision (MTTC) metric, described in \cite{lit:risk} to quantify the risk, $r$.
\paragraph*{Modified TTC} 
MTTC at an instant $t$ is defined as the time that remains until a collision between two vehicles would have occurred if the collision course, speed difference, and current accelerations are maintained.\\
\cite{lit:risk} Notes that the drawback of MTTC is that it assumes that the vehicles are on the same collision course which might not always be the case, thus, looming is introduced. In this work the PLG allows us to circumvent this drawback as we can plan and predict all vehicle paths and their associated probabilities using (\ref{eq:path plan}).

We then simulate driver behaviour by running (\ref{eq:path plan}) to guide the vehicles to their target destinations and (\ref{eq:action}) to allow the vehicles to respond to risk due to BVs. To generate corner case scenarios we learn a new policy that interacts with surrounding agents more aggressively, hence, resulting in an increased crash rate in our simulated data sets. We modify the action-generating part of our model, (\ref{eq:path plan}), since this is the part that responds to the surrounding agents. Therefore, we replace the original policy, $p(a_k, L_k~|~\mathbf{r}_k)$, with the learnt, more aggressive, policy $\pi_p(a_k, L_k~|~\mathbf{r}_k)$.

The amount by which the original policy, $p(a_k,L_k~|~\mathbf{r}_k)$, is modified needs to be constrained via some means to prevent the algorithm from learning unrealistic policy distributions. To this end, we use a Proximal Policy Optimisation (PPO) actor-critic style algorithm \cite{lit:PPO} to optimise (\ref{eq:action}). The loss function used is the clipped surrogate loss function given by:
\begin{equation} \label{eq:loss function}
    L^{CLIP}(\phi) = \mathbb{E}_k \left[\min(r_k(\phi) \hat{A}_k, \text{clip}(r_k(\phi), 1-\epsilon, 1+\epsilon)\hat{A}_k) \right]
\end{equation}
This will ensure that we do not stray too far from the original policy, $p(a,L~|~\mathbf{r})$, by clipping the gradients if they become too large. $\hat{A}_k$ is the advantage estimate at each time step $k$ as described in \cite{lit:PPO}. We take the reward assigned to the algorithm at each iteration as the risk experienced by the vehicle at that particular time. The reward (risk) is therefore taken as the inverse of the MTTC:
\begin{equation}
    r = \frac{1}{MTTC}
\end{equation}
In doing so we can receive a large reward for near-crash events (low MTTC) which quickly tapers off as MTTC increases and the possibility of collision decreases. If two vehicles collide they will have a time to collision of zero. However, the simulation is terminated once a crash event occurs, therefore, a divide by zero case is never encountered. Furthermore, we note that since we are working in a discretised state space MTTC will rarely take extremely low values which causes $r$ to grow extremely large.

\section{Experimental Results}
\label{sec:experiments}

In this section, we present quantitative and qualitative results for the generation of corner cases using one of the learnt PLGs.  


\subsection{PLG Generation}
To generate a PLG and learn the corresponding vehicle behaviour we require a data set $\mathcal{D}$ to contain the following information about each vehicle: two-dimensional spatial coordinates, speed, and acceleration. However, we can significantly improve upon our knowledge of vehicle behaviour if a lane ID is associated with each vehicle's position for each data point. Note that we do not need the direction of travel as this will be encoded into the adjacency matrix of the PLG. To show the generality of our approach and its utility in a diverse set of road networks including highways, intersections, and roundabouts we have selected the following datasets: NGSIM~\cite{dataset:NGSIM} and  rounD~\cite{dataset:rounDdataset} and shown the example PLG generated in \cite{GitHub:plg-generation}.

\subsection{Corner Case Generation: Quantitative Results}
In the following we are discussing the corner case generation using the PLG learnt from NGSIM~\cite{dataset:NGSIM} (Fig.~\ref{fig:PLG}). 

Post discretisation, there are a total of 1,692 unique trajectories and 263,410 different states in the data set $\mathcal{D}$. From this, we extract all high-risk states (see Section~\ref{sec:cornercase}) and use these as seed states to initialise our simulations. We compare our corner case generation rate, $r_{cc}$, to that achieved in \cite{lit:corner_case_gen} and the baseline number of corner cases contained in the data set $\mathcal{D}$ as a result of the discretisation in Table~\ref{tab:ccrate}. The NGSIM data set itself only includes a very small number of corner cases ($r_\text{cc}=0.0056$). Using our proposed approach based on PLGs combined with our definition for risk we achieve a corner case rate of $r_\text{cc}=0.416$. It also outperforms the RL-based approach in \cite{lit:corner_case_gen} which does not use PLGs.

\begin{table}
\begin{center}
\caption{Corner case rate in NGSIM \cite{dataset:NGSIM}}
\label{tab:ccrate}
\begin{tabular}{ll}
\toprule
\textbf{Approach} & \textbf{Corner case rate ($r_\text{cc}$)}\\
\midrule
   
     PLG + observed traffic data in  $\mathcal{D}$ (baseline) & 0.0056 \\
     PLG + RL-based corner case generation & \textbf{0.416}\textbf{}\\\midrule
     RL-based corner case generation\cite{lit:corner_case_gen} &  0.3 \\
 \bottomrule
\end{tabular}
\end{center}
\end{table}

We classify the generated corner cases into three distinct cases which are defined below. The classifications are made by considering vehicle data up to $T$ seconds before the collision.

\begin{itemize}
\item \emph{Case 1:} No lane changes. The scenario corresponds to a one-dimensional case (the dimension we refer to in this case is along the length of the lane).
\item \emph{Case 2:} One lane change by either vehicle.
\item \emph{Case 3:} Two (or more) lane changes in total.
\end{itemize}

The generated corner cases are distributed among \emph{Case 1}, \emph{Case 2} and \emph{Case 3} in the proportions $0.807$, $0.188$ and $0.005$ respectively. Additionally, of the corner case scenarios, $76$\% occurred at or around an intersection. Although 0.5\% of constitutes a small proportion of the overall results for two-lane change scenarios, the method can be adapted quite easily in order to facilitate a greater number of two-lane change scenarios. This is however beyond the scope of this work and thus is left for future work


 

\subsection{Corner Case Generation: Qualitative Examples}
In this section, we discuss examples of each of the three types of corner cases. First, we provide qualitative descriptions of each case and extract explanations from the PLGs. The discussed corner cases are depicted in Fig.~\ref{fig:cc cases}.



\paragraph*{Case 1 (Fig.~\ref{subfig:cc_case_1})} A one-dimensional crash event between the AV (in black) and a slow-moving BV (in green but plotted behind the black vehicle, therefore, it cannot be seen). The black vehicle (ID 834) was the faster-moving vehicle and caused a rear-end collision with vehicle 845. Looking at the time stamps on the path taken by vehicle 834, we see that it maintained a relatively constant velocity as the difference between the times it transitioned nodes remains about the same. The speed difference between the two vehicles was relatively low, hence, due to the more aggressive action policy the vehicles were operating under, lower magnitude actions were taken to minimise risk.

\paragraph*{Case 2 (Fig.~\ref{subfig:cc_case_2})} A two-dimensional crash event between vehicle 1322 and 1327. Vehicle 1327 performs a left lane change which results in a side impact collision as vehicle 1322 was passing through the same point in space at the time of the lane change. Vehicle 1327 attempted to change lanes to perform a left turn at the intersection just ahead of the collision point. We deduce this by studying (\ref{eq:path plan}), where we can notice that most paths taken by vehicles in the left two lanes contain a left turn at the stated intersection. Furthermore, looking at the nodes ahead of vehicle 1327, before it attempted the lane change manoeuvre, we notice that the probability of performing a lane change further up the road reduces substantially. We can see that the lane change edges in the PLG are considerably lighter, indicating that lane changes further up the road were far less common. The vehicle’s increased tolerance to risk coupled with the low probability of changing lanes in the future is the trigger for the atypical behaviour. Hence, explaining the attempted lane change manoeuvre resulting in the collision event.

\paragraph*{Case 3 (Fig.~\ref{subfig:cc_case_3})} A two-dimensional crash event between vehicles 841 and 843. Both vehicles attempted a lane change manoeuvre attempting to access the same lane simultaneously at the same point in space. Again, by studying (\ref{eq:path plan}), we deduce that vehicle 841 attempted a left lane change so it could perform a left turn at the intersection at the end of the road. Vehicle 841 attempted a right lane change, so it could either access a lane further to the right in the near future and perform a right turn, or continue straight ahead at the intersection. Looking at the PLG edges we see that vehicle 843 had less possibility of performing a right lane change manoeuvre further down the road, hence, its risky attempt to change lanes here. We note, however, that such a collision could have been avoided. The PLG shows that vehicle 841 had other opportunities to transition lanes further down the road, however, due to the aggressive action policy it attempted the lane change in a riskier state, ultimately resulting in the bumper-to-bumper collision event.

\subsection{Summary}

The discretisation of the search space for vehicle trajectories and path planning allows us to model drivers using a set of low-dimensional, more explainable models as opposed to a complex black box type model. Consider now, the incorporation of the traffic protocols described in \cite{lit:graph_representation}. Over the course of a simulation, this results in a dynamically changing PLG where nodes and edges would be accessible to a vehicle traversing the PLG depending on whether a background vehicle is occupying this node. Whether or not a node is accessible directly affects the trajectory of a vehicle in a predictable and explainable way. This is a result of the fact that trajectories are generated directly using the probabilistic adjacency matrix of our PLG. To gain insight into a chosen trajectory of a vehicle would simply mean evaluating the state of the adjacency matrix at that point in time.

Vehicles may traverse the PLG from start to finish even when lane IDs are omitted from the nodes. However, if we can associate a lane ID to each node, we can extract the scene graphs described in \cite{lit:scene_graphs} directly from our PLG. In doing so, we can provide an increased degree of explainability to each scene at every point in time throughout the simulation.

In \cite{GitHub:plg-generation}, we provide videos that visualise some corner case data we have generated. We also show an example of a generated scenario that has been smoothed using a moving average filter in order to convert the sharp paths within the PLG into smoother and more realistic trajectories.

\section{Conclusion}
\label{sec:conclusion}

\subsection{Discussion \& Future Work}
Some limitations of our modelling process in this paper are as follows:
\begin{itemize}
    \item The model did not account for veering within a lane. This is something that can be easily rectified by introducing some stochasticity to each node.
    \item Lack of an optimal method to calculate minimum node distance $R$.
    \item When continuous paths in the data set, $\mathcal{D}$, are discretised, the discrete paths can exhibit a meandering motion between lanes during vehicle lane changes. This undesired motion can be rectified by using a more sophisticated method to discretise the continuous paths. For simplicity, we used the naive approach of assigning a vehicle to the node it was closest to without placing any statistical significance on the decision.
    \item Data that is spread across a large region, i.e., a small amount of data but in many different locations, limits the variety of different vehicle paths we can generate. Whereas a larger amount of data in a few locations allows us to generate a richer variety of paths a vehicle can take. In our case the NGSIM dataset was sufficient for our method to generate successful results, however, repeating these experiments with more data would yield an improved model of driver behaviour. 
\end{itemize}

We note the model can be further developed to generate improved and more complex vehicle motion. Consider instead of a single path, generating a tree of possible paths, branching off at each node where there are multiple possible future nodes to visualise all future paths for a vehicle. Furthermore, note that the adjacency matrix for the PLG is sparse, such matrices have been known to be highly compressible in graph theory, hence, providing an efficient software implementation of the methodology. We extend this concept by noticing that layers of complexity can be added to the overall model of the environment without increasing the dimensionality of the action-generating model. One can imagine a set of adjacency matrices where each matrix provides additional information about the environment and road conditions.  In this paper we have focused on different single-lane and lane-changing scenarios, however, the method may also be extended to different scenarios.

Although our action model stems from human driving data, as long as human drivers outnumber AVs, all corner cases are realistic but atypical, providing novel cases for AV testing. 
Natural next steps would be to test state-of-the-art vehicle models within the framework we've developed and, further evaluating how realistic the generated scenarios are by testing if the crashes are avoidable by an AV.

\subsection{Conclusion}
Overall, we have introduced Probabilistic Lane Graphs (PLGs) as a novel and powerful methodology in the context of corner case generation in autonomous driving. Using the PLG we can provide highly intuitive explanations to generated corner cases as well as produce useful visual descriptions of the scenario and environment. We also introduce our novel parallel framework using the PLG to separate the vehicle path planning and action generation into two mostly independent models. In doing so, we further increase the explainability of our model since we are now working with two low-dimensional models, rather than a single high-dimensional model. We then use reinforcement learning to modify the action-generating part of our model to learn a more aggressive policy which increases the rate of corner case events. We test our methodology on \cite{dataset:NGSIM} and provide a variety of visualisations of generated corner case scenarios.







\end{document}